\documentclass[letterpaper]{article} %
\usepackage[preprint]{aaai2027}  %
\usepackage[hyphens]{url}  %
\usepackage{graphicx} %
\usepackage{natbib}  %
\usepackage{caption} %
\usepackage{algorithm}
\usepackage{algorithmic}
\usepackage{amsmath}
\usepackage{amssymb}
\usepackage[inline]{enumitem}
\usepackage{xspace}
\usepackage{dsfont}
\usepackage{amsthm}

\newcommand{\nesywm}{\text{NeSy-WM}\xspace}
\newcommand{\nesywms}{\text{NeSy-WMs}\xspace}

\newlist{assumptions}{enumerate*}{1}

\setlist[assumptions]{label=\textbf{(A\arabic*)}}

\newcommand{\mctsresult}[3]{%
  \shortstack{%
    $#1\!\pm\!#2$\\
    $\Delta=#3$%
  }%
}

\newcommand{\mctsnosuccess}[2]{%
  \shortstack{%
    $#1\!\pm\!#2$\\
    $\Delta=\text{\phantom{0.}--\phantom{0}}$%
  }%
}

\newcommand{\indicator}[1]{\ensuremath{\mathds{1}_{\left\{#1\right\}}}\xspace}

\definecolor{mango_tango}{HTML}{F9844A}

\newtheorem{example}{Example}[section]

\usepackage{newfloat}
\usepackage{listings}
\DeclareCaptionStyle{ruled}{labelfont=normalfont,labelsep=colon,strut=off} %
\floatstyle{ruled}
\newfloat{listing}{tb}{lst}{}
\floatname{listing}{Listing}

\usepackage{booktabs}

\title{Towards Zero-Shot Task Transfer with Neurosymbolic World Models}

\author{
    Isidoro Tamassia,
    Lennert De Smet,
    Giuseppe Marra
}

\affiliations{
    KU Leuven, Department of Computer Science, Belgium\\
    \{isidoro.tamassia, lennert.desmet, giuseppe.marra\}@kuleuven.be
    
}

\begin{document}

\maketitle

\begin{abstract}
State-of-the-art model-based reinforcement learning methods learn neural world models that allow policy improvement by planning in a latent space, without assumptions on the structure of the underlying environment. While expressive, these models are generally task-dependent: they learn uninterpretable latent representations that are tied to the training task and thus hard to generalize to new tasks. In this work, we present a novel world model formulation where the reward prediction only depends on a subset of structured, symbolic components of the whole latent state. Decoupling observation reconstruction and reward prediction allows us to learn world models that can adapt zero-shot, i.e. without further environment interactions, to new reward functions defined over the same symbolic state space. We discuss the main advantages and challenges of learning these neurosymbolic world models and demonstrate the strong generalisation properties of our approach over purely neural methods. 
\end{abstract}

\section{Introduction}
World models are motivated by the idea that agents can learn compact internal representations of their environments from experience and use them to guide decision-making~\cite{ha2018recurrent}.
In reinforcement learning (RL), popular methods typically learn latent representations of states, dynamics, and reward functions directly from high-dimensional observations such as sequences of images~\cite{muzero, hafner2019learning, hafnerdreamv1, hafnermasteringv2, hafner2025mastering, tdmpc, tdmpc2}.
These learned representations can  be used to optimise behaviour in several ways, including model-predictive control~\cite{tdmpc, tdmpc2}, Monte-Carlo Tree Search~\cite{muzero}, or actor-critic learning on imagined trajectories~\cite{hafnerdreamv1, hafnermasteringv2, hafner2025mastering}.

Despite their ability to learn compact representations of environment dynamics, RL world models are not straightforwardly reusable across tasks.
A particularly important case is \textit{task transfer under shared dynamics}: the environment and its transition structure remain the same, but the reward function changes.
This setting arises naturally in domains where the same environment supports multiple objectives, such as reaching different goals, placing objects in different configurations, or avoiding different hazards. Current RL world models are poorly suited to this setting because they typically learn a reward predictor jointly with the latent state and dynamics model.
As a result, the learned representations may depend fully~\cite{muzero,tdmpc,tdmpc2} or partially~\cite{hafnerdreamv1,hafnermasteringv2,hafner2025mastering} on the rewards observed during training.
When the task changes, the learned reward predictor remains tied to the training objective and is therefore not aligned with the new reward function, even though the underlying dynamics are unchanged.
At the same time, the latent states learned by these models do not provide an interpretable interface on which a new reward function can be specified directly.
Consequently, reusing a task-dependent world model for a new task typically requires learning or finetuning a new reward predictor \emph{using data from the test environment}, or by \textit{relabeling training trajectories} with the downstream task rewards~\cite{sekar2020planning}.

Unlike transition dynamics, reward functions often encode the desiderata of a user or task designer rather than an intrinsic property of the environment.
Hence, they are frequently specified in terms of high-level semantic abstractions of the state, such as an agent reaching a target location, objects satisfying a desired configuration, or unsafe states being avoided~\cite{icarte2022reward}.
When a reward function is  given, replacing it with a learned predictor can unnecessarily entangle the task objective with the latent dynamics model.

However, directly using an explicit reward function requires the variables on which it depends to be available to the model. If these symbolic properties were available as part of the model state, 
the same learned model could be reused for new tasks whose reward functions are defined over the same properties, even when the reward specification itself changes.
Crucially, exposing such symbolic properties should not require replacing learned latent dynamics with a fully symbolic model.
The symbolic component only needs to capture the reward-relevant aspects of the state, while the remaining information required for control can remain encoded in latent representations learned from raw, subsymbolic observations.

We instantiate this principle with \textbf{Neurosymbolic World Models (\nesywms)}, a new class of world models that combine latent dynamics learning with an explicit symbolic interface for reward evaluation.
\nesywms build on Recurrent State-Space Models (RSSMs)~\cite{hafner2019learning}, which learn compact latent dynamics from high-dimensional observations.
In addition, a \nesywm predicts a set of symbolic state properties that are chosen to capture the variables on which task rewards depend.
Rewards are then predicted only through these symbolic properties, rather than directly from the latent state.
This design disentangles latent modeling of environment dynamics from symbolic, interpretable reward prediction.
Importantly, the symbolic state does not need to provide a complete symbolic description of the environment.
It only needs to expose the reward-relevant properties, leaving the remaining information needed for prediction and control in the learned latent representation.

Our contributions are as follows.
\begin{enumerate*}[label=\textbf{(\arabic*)}]
    \item We formalise \nesywms and their use for task transfer under shared dynamics, where new reward functions are defined over the same symbolic state properties as the training task.
    \item We propose different symbolic supervision regimes for \nesywms 
    and evaluate their training performance and sample-efficiency against the state-of-the-art DreamerV3~\cite{hafner2025mastering}.
    \item We evaluate the zero-shot adaptation of \nesywms to new test-time tasks in two different settings: one where pure imagination planning is performed to solve the task (no further learning), and one where the model is finetuned in imagination.
    \item We provide an open-source and highly configurable PyTorch~\cite{paszke2019pytorch} implementation of \nesywms.
\end{enumerate*}

\section{Related Work}
\label{sec: related_work}
Our work lies at the intersection of world models for model-based RL (MBRL) and probabilistic neurosymbolic learning.

World models aim to predict, explicitly or implicitly, how an environment (the world) evolves under actions. While reward-free world models are typically trained from observations alone through predictive or generative objectives~\cite{assran2023self, zhou2025dino, micheli2023transformers}, several task-dependent world models used in MBRL instead aim for optimal control by learning latent dynamics directly through reward prediction, value prediction, and policy improvement objectives~\cite{muzero, danihelka2022policy, tdmpc, tdmpc2}. 
Generative world models~\cite{ha2018recurrent,alonso2024diffusion,hafnerdreamv1,hafnermasteringv2,hafner2025mastering} additionally shape their representations by \textit{reconstructing} observations or sensory inputs.
In particular, Recurrent State-Space Models (RSSMs)~\cite{hafner2019learning} can successfully model high-dimensional, partially observable environments and form the backbone of the Dreamer family of algorithms~\cite{hafnerdreamv1,hafnermasteringv2,hafner2025mastering}.
Moreover, their probabilistic formulation is naturally compatible with probabilistic neurosymbolic methods.

Neurosymbolic AI~\cite{garcez2023neurosymbolic,marra2024statistical} integrates symbolic knowledge and reasoning into machine learning models.
Various probabilistic neurosymbolic models~\cite{manhaeve2018deepproblog,yang2020neurasp} now integrate generative models to constrain generations
and perform test-time interventions.
They mostly use variational autoencoders~\cite{kingma2014auto} with a static~\cite{misino2022vael,de2023neural} or dynamic~\cite{desmet2025relational} encoding space combining latent and symbolic features.
Their probabilistic nature is not only used for compatibility with generative models, but also to enable end-to-end training via probabilistic inference.
A number of neurosymbolic approaches to world modeling have recently been proposed.
\citet{cano2025neurosymbolic} learn reusable neural world-model primitives
from offline data and assembles them into a finite-state-machine world model to facilitate control across environments.
\citet{sehgalneurosymbolic}   introduce an object-centric neurosymbolic world model that augments object representations with symbolic attributes to improve compositional generalisation in dynamics prediction.
In both cases, the primary goal is to leverage symbolic structure to improve modeling and generalisation of environment dynamics, rather than our goal of enabling adaptation to new tasks under shared dynamics.

Finally, our work is related to goal-conditioned RL (GCRL), which learns policies or value functions conditioned on a goal representation to solve a family of tasks~\cite{schaul2015universal, andrychowicz2017hindsight, liu2022goal}. GCRL typically amortizes adaptation by \textit{jointly training on many goals} and executing a goal-conditioned policy at test time. In contrast, we learn a NeSy-WM  \textit{on a single training task} and perform the desired task adaptation in the imagination of the world model, without the need to pre-train in a multi-task setting.
\section{Methodology}
\label{sec: methodology}
\subsection{Overview}
\label{subsec: overview}

Our goal is \emph{task transfer} under two main assumptions.
\begin{assumptions}
    \item Training and test reward functions are explicitly defined in terms of \emph{a limited number} of high-level abstractions of the state.
    Such abstractions will be called the symbolic properties of the state \emph{and the vocabulary of such abstractions is assumed to be given by the user}.
    \label{assumption1}
    \item The underlying dynamics of all tasks is the same.
    \label{assumption2}
\end{assumptions}
For instance, consider a navigation environment 
where the training task is to reach a fixed goal.
Whether this task has been achieved only depends on the location of the agent.
No other properties of the environment, e.g. agent orientation, matter for assessing task-achievement.
A possible test task would be to shift the goal to any other position in the map;
the reward function still depends only on the agent location, despite inducing a task that is completely different from the training task.

Current world models \emph{learn purely latent representations} of the state which cannot directly exploit user-specified reward functions, even when the test tasks are naturally defined on symbolic properties.
As a result, a new reward predictor has to be learned, e.g. by manually relabeling replayed observations~\cite{sekar2020planning}.

\begin{figure}[t]
    \centering
    \includegraphics[width=0.8\linewidth]{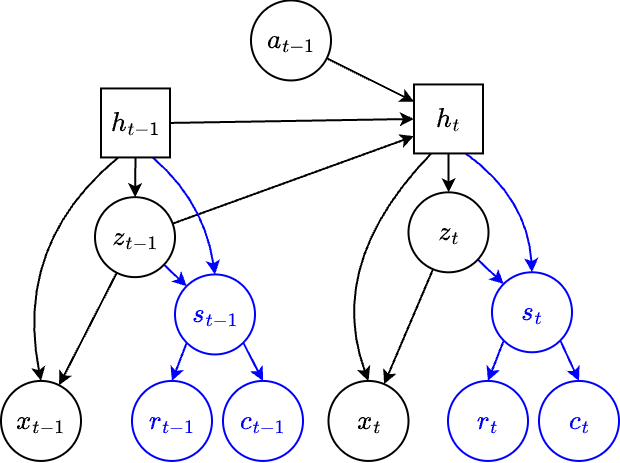}
    \caption{Overview of \nesywms. The symbolic-bottlenecked reward and continue predictors are shown in blue; the rest follows the RSSM structure of~\cite{hafner2019learning}. At test time, new reward and continue functions can be defined over the same symbols to enable zero-shot planning or imagination training of a new actor-critic.}
    \label{fig: nesy-wm-graphic}
\end{figure}

\paragraph{\nesywms avoid learning new reward functions without harming representation learning.}
They instead make the symbolic properties an explicit part of the world model.
The symbolic properties form a bottleneck for the reward prediction to facilitate substitution of any other reward function defined on the same properties.
More specifically, the model augments DreamerV3~\cite{hafner2025mastering} with \emph{neurosymbolic reward and continue predictors} (Figure~\ref{fig: nesy-wm-graphic}).
\begin{align*}
\textbf{Sequence model:} \ & h_t = f_\phi(z_{t-1}, h_{t-1}, a_{t-1}) \\
\textbf{Encoder:} \ & z_t \sim q_\phi(z_t \mid h_t, x_t) \\
\textbf{Dynamics predictor:} \ & \hat{z}_t \sim p_\phi(\hat{z}_t \mid h_t) \\
\textbf{NeSy reward predictor:} \ & \hat{r}_t \sim p^{\text{sym}}_\phi(\hat{r}_t \mid z_{t}, h_{t}) \\
\textbf{NeSy continue predictor:} \ & \hat{c}_t \sim p^{\text{sym}}_\phi(\hat{c}_t \mid z_{t}, h_{t}) \\
\textbf{Decoder:} \ & \hat{x}_t \sim p_\phi(\hat{x}_t \mid z_{t}, h_{t})
\end{align*}
Here, $x_t$ denotes the observation at time $t$, $a_t$ the action, $h_t$ the deterministic recurrent hidden state, $z_t$ the stochastic latent state. The "continue" predictor models the probability that the episode continues at the next step, i.e. that the current state is non-terminal. 
The function $f_\phi$ is the deterministic recurrent transition of the RSSM.
Next, we specify the neurosymbolic predictors and how to compute their probabilities.

\paragraph{Neurosymbolic predictors combine neural parametrisations with functions on symbols.}
The neurosymbolic reward predictor first neurally predicts a distribution
$
p_{\phi}(s_t \mid z_t, h_t)
$
over the symbolic properties using a simple MLP.
The symbolic properties are assumed to be the only input to a given reward function
$
p^{sym}(r_t \mid s_t)
$
(Assumption~\ref{assumption1}).
The uncertainty on $s_t$ parametrised by
$
p_{\phi}(s_t \mid z_t, h_t)
$
induces a probability distribution over the rewards $r_t$.
Indeed, the distribution $p_\phi^\text{sym}(r_t \mid z_t, h_t)$ is given by marginalising over the symbolic states $s_t$ as
\begin{align}
    \label{eq:wmc}
    &\sum_{s_t}
    p^{sym}(r_t \mid s_t)
    p_{\phi}(s_t \mid z_t, h_t)
    \\
    \label{eq:wmi}
    &\text{or}
    \qquad
    \int
    p^{sym}(r_t \mid s_t)
    p_{\phi}(s_t \mid z_t, h_t)
    \ \mathrm{d}s_t
    ,
\end{align}
for discrete or continuous symbolic properties, respectively.
Equations \ref{eq:wmc} and \ref{eq:wmi} also explain the generalisation ability of \nesywms;
 $p^{sym}(r_t \mid s_t)$ can simply be replaced with a new reward function.
The same holds for the neurosymbolic continue predictor, which can be defined analogously. 

\begin{example}[Discrete Navigation]
    \label{ex:minigrid}
    Consider a discrete grid world where the symbolic property $s_t = (x_t, y_t)$ models the discrete coordinates of the player at time step $t$.
    The reward function
    $
    p^{sym}(r_t \mid s_t)
    $
    takes the coordinates $s_t$ as input and returns $1$ if $s_t$ is the goal  $(x_g, y_g)$ and $0$ otherwise.
    If we model the distribution over the symbolic properties $p_\phi(s_t \mid z_t, h_t)$ as two independent categorical distributions, then $p_\phi^{\mathrm{sym}}(r_t = 1 \mid z_t, h_t)$ (Equation~\ref{eq:wmc}) reduces to 
    \[
    \Pr_\phi(x_t=x_g \mid z_t,h_t)\cdot\Pr_\phi(y_t=y_g \mid z_t,h_t).
    \]
\end{example}

\begin{example}[Continuous Navigation]
    \label{ex:miniworld}
    Consider a continuous grid world that has a continuous symbolic state $s_t = (x_t, y_t) \in \mathbb{R}\times\mathbb{R}$.
    Here, a goal is reached when the symbolic state $s_t$ lies in a square target area
    $
    \left\{
    (x, y)
    \mid
    a_x^{-} \leq x \leq a_x^{+},
    a_y^{-} \leq y \leq a_y^{+},
    \right\}
    $
    .
    If we model the distribution over the symbolic properties $p_{\phi}(s_t \mid z_t, h_t)$ as two independent Gaussian distributions
    \begin{align*}
        x_t &\mid z_t,h_t \sim \mathcal N(\mu_x(z_t,h_t),\sigma_x^2(z_t,h_t)),
        \\
        y_t &\mid z_t,h_t \sim \mathcal N(\mu_y(z_t,h_t),\sigma_y^2(z_t,h_t)),
    \end{align*}
    then $p_\phi^\text{sym}(r_t=1 \mid z_t,h_t)$ (Equation~\ref{eq:wmi}) is given by
    \begin{align*}
        \Pr_\phi(a_x^- < x_t < a_x^+)\cdot
        \Pr_\phi(a_y^- < y_t < a_y^+),
    \end{align*}
    where $\Pr_\phi(a_x^- < x_t < a_x^+)$ is computable using the Gaussian CDF (supplementary material).
\end{example}
In the relevant case where the reward function checks whether $s_t$ satisfies logical properties, computing Equations \ref{eq:wmc} and \ref{eq:wmi} reduces to the standard inference tasks in probabilistic neurosymbolic AI:
Weighted Model Counting (WMC)~\cite{chavira2008probabilistic} for discrete $s_t$ or Weighted Model Integration (WMI)~\cite{belle2015probabilistic} for continuous $s_t$.
WMC and WMI have highly-optimised inference strategies~\cite{chavira2008probabilistic,zeng2020scaling,maene2025klay} to better handle the intractability of computing Equations \ref{eq:wmc} and \ref{eq:wmi} \cite{roth1996hardness}. 

\subsection{Model Learning}
\label{subsec: model_learning}
Our model-learning setup differs from DreamerV3 in two ways.
\begin{enumerate*}[label=\textbf{(\arabic*)}]
    \item The neurosymbolic predictors $p^{\text{sym}}_\phi(r_t \mid z_t, h_t)$ and $p^{\text{sym}}_\phi(c_t \mid z_t, h_t)$ replace the  standard MLPs and
    \item symbolic supervision losses that can be added to ensure aligned symbolic representations. 
\end{enumerate*}

\paragraph{Standard Loss}
Given a sequence of observations $x_{1:T}$, actions $a_{1:T}$, rewards $r_{1:T}$ and continuation flags $c_{1:T}$, we minimize the overall world model loss
\begin{align*}
\label{eq:world-model-loss}
\mathcal L_{\mathrm W}(\phi)
&=
\mathbb E_{q_\phi}\Bigg[
\sum_{t=1}^{T}
\Big(
    \beta_{\mathrm{pred}}\mathcal L_{\mathrm{pred}}^t(\phi)
    + \beta_{\mathrm{dyn}}\mathcal L_{\mathrm{dyn}}^t(\phi)
\\[-0.2em]
&\hspace{4.8em}
    + \beta_{\mathrm{rep}}\mathcal L_{\mathrm{rep}}^t(\phi)
    + \beta_{\mathrm{rec}}\mathcal L_{\mathrm{rec}}^t(\phi)
\Big)
\Bigg],
\nonumber
\end{align*}
where $ \beta_{\mathrm{pred}}, \beta_{\mathrm{dyn}},\beta_{\mathrm{rep}}, \beta_\text{rec}$ are fixed hyperparameters and the individual losses are defined as
\begin{align*}
\mathcal L_\text{rec}^t(\phi) &= -\log p_{\phi}(x_t \mid z_{t}, h_{t}), \\
\mathcal{L}^t_{\mathrm{pred}}(\phi) &= -\log{ p^\text{sym}_{\phi}(r_t \mid z_{t}, h_{t})} - \log p^\text{sym}_{\phi}(c_t \mid z_t,h_t), \\
\mathcal{L}^t_{\mathrm{dyn}}(\phi) &= \max\bigl(1,\text{KL}[\text{sg}( q_{\phi}(z_t \mid h_t,x_t)) \, || \, p_{\phi}(z_t \mid h_t)]), \\
\mathcal{L}^t_{\mathrm{rep}}(\phi) &= \max(1,\text{KL}[ q_{\phi}(z_t \mid h_t,x_t) \, || \,\text{sg}(p_{\phi}(z_t \mid h_t))]),
\end{align*}
where $\text{sg}(\cdot)$ is the stop-gradient operator.
The dynamics loss $\mathcal{L}_\text{dyn}^t(\phi)$  and representation loss $\mathcal{L}_\text{rep}^t(\phi)$ optimise the transitions via a KL divergence between prior and posterior.
Reconstruction loss $\mathcal{L}_\text{rec}^t(\phi)$ and prediction loss $\mathcal{L}_\text{pred}^t(\phi)$ ensure accurate observations and reward/continuation signals.

\paragraph{Neurosymbolic predictors are differentiable.}
The neurosymbolic reward and continuation predictors are computed exactly via Equation \ref{eq:wmc} or \ref{eq:wmi}.
Exact neurosymbolic inference remains end-to-end differentiable~\cite{manhaeve2018deepproblog} to propagate gradients from $s_t$ to $z_t$ and $h_t$.
Consequently, the neurosymbolic $p^\text{sym}_{\phi}(r_t \mid z_{t}, h_{t})$ and $p^\text{sym}_{\phi}(c_t \mid z_t,h_t)$ are drop-in replacements for the MLPs used by DreamerV3.

\paragraph{Symbolic supervision losses make symbolic states identifiable.}
Solely relying on task reward prediction as a source of distant supervision for learning symbolic states can be insufficient to learn aligned representations~\cite{marconato2023not}.
Identifiability of the symbols impacts task generalisation;
if the symbols are learned wrongly, symbolic states may be assigned wrong rewards at test-time.
For this reason,  we consider and test three different supervision schemes \textbf{(S)}. 
\begin{enumerate}[
    label=\textbf{(S\arabic*)},
    ref=S\arabic*,
    leftmargin=0pt,
    labelsep=0.5em,
    labelwidth=0pt,
    itemindent=*,
    align=left,
    itemsep=0.75em
]

\item \label{S1} \textbf{Full supervision.}
Direct symbolic supervision on \emph{all} the states visited throughout world model training. This requires the agent to directly observe the symbolic properties $s_{1:T}$. The per-step symbolic supervision loss is the negative log-likelihood of the ground-truth symbol
\begin{equation*}
    \mathcal L^t_\text{sym}(\phi)
    =
    -\log p_\phi(s_t \mid z_t, h_t).
\end{equation*}
This kind of supervision can be seen as a form of privileged-information training~\cite{lambrechts2024informed, huang2025pigdreamer}. However, such approaches either provide the full Markovian state during training, or a surrogate that is sufficient for reconstructing image observations~\cite{lambrechts2024informed}. This is a stronger assumption than observing the symbolic properties considered in our work, which neither constitute a Markovian state representation nor suffice to reconstruct image observations.

\item \label{S2} \textbf{Partial supervision.}
A weaker assumption is to supervise only a subset $\mathcal{S}'$ of all symbolic states. This limits the generalisation of \nesywms to novel tasks involving relevant symbolic states, e.g. potential goal states, in $\mathcal{S}'$. Concretely, partial supervision adds the log-likelihood of the ground-truth symbols only when a state in $\mathcal{S}'$ is visited:
\begin{equation*}
    \mathcal L^t_\text{sym}(\phi)
    =
    -
    \indicator{s_t \in \mathcal{S}'}
    \cdot
    \log p_\phi(s_t \mid z_t, h_t).
\end{equation*}
Intuitively, this corresponds to having sensors in limited parts of the environment. A potential issue is that the predictor may \textit{collapse} to always predicting supervised symbols; after a reward intervention, a state incorrectly
mapped to a rewarded supervised symbol may produce a spurious reward and cause
transfer failure.
A simple solution inspired by~\citet{marconato2024bears} is to add an entropy regularisation term to discourage overly confident  predictions. For a symbolic state composed of $M$ categorical variables, we minimise
\begin{equation*}
    \mathcal L^t_{\mathrm{ent}}(\phi)
    =
    -\beta_{\mathrm{ent}}
    \frac{1}{M}
    \sum_{j=1}^{M}
    \frac{
        \mathcal H\!\left[
            p_\phi(s_t^{(j)} \mid z_t,h_t)
        \right]
    }{
        \log |\mathcal S_j|
    },
\end{equation*}
where $\mathcal H$ denotes Shannon entropy, $\mathcal S_j$ is the domain of the $j$-th symbolic variable, and $\beta_{\mathrm{ent}}$ controls the regularisation.

\item \label{S3} \textbf{No supervision.}
When no supervision is provided, the symbolic predictors can learn any mapping as long as it suffices to solve the training task. Nonetheless, we still consider this setting because the imposed symbolic structure may still help stabilise training and facilitate downstream finetuning.
\end{enumerate}

\subsection{Test-Time Interventions and Generalisation}
\label{sec:ttime-int-gen}
At training time, control follows the actor-critic scheme of DreamerV3, summarised in the supplementary material.
Once a \nesywm is trained, it can be adapted to different tasks \emph{without access to the test environments}.
To do so, it is sufficient to replace the training predictors $p^{sym}(r_t \mid s_t)$ and $p^{sym}(c_t \mid s_t)$ in Equation \ref{eq:wmc} or \ref{eq:wmi} with the given test predictors.
This requires new tasks to be defined over the same symbols (Assumption~\ref{assumption1}) and assumes fixed dynamics (Assumption~\ref{assumption2}).
Given the replaced reward function, there are two options \textbf{(O)} to adapt a \nesywm to a novel test task.
In both cases, \emph{there is no interaction with the test environment}.
\begin{enumerate}[
 label=\textbf{(O\arabic*)},
    ref=O\arabic*,
    leftmargin=0pt,
    labelsep=0.5em,
    labelwidth=0pt,
    itemindent=*,
    align=left,
    itemsep=0.75em
]

\item \label{O1} \textbf{Adaptation by pure planning.}
Monte-Carlo Tree Search (MCTS)~\cite{coulom2006efficient,kocsis2006bandit} can be employed to plan at test time using the learned model and the new reward/continue predictors \emph{without any further learning}. We stress that this is not possible for purely neural models such as Dreamer, since they cannot replace the previously learned reward/continue predictors without a finetuning phase. While we experiment on environments with discrete action spaces, \nesywms can also be used for planning in continuous action spaces using compatible algorithms, such as the Cross-Entropy Method (CEM)~\cite{rubinstein1997optimization} or appropriate extensions of MCTS~\cite{couetoux2013monte}.

\item \label{O2} \textbf{Adaptation by imagination finetuning.}
NeSy-WMs can be finetuned on a new task entirely in the world model's imagination \emph{without any further interaction with the environment}. By manually replacing the predictors $p^{\mathrm{sym}}(r_t \mid s_t)$ and $p^{\mathrm{sym}}(c_t \mid s_t)$ as described above, a new actor-critic can be trained without collecting new task-specific data or performing the time-consuming replay-buffer relabeling required by previous Dreamer adaptation approaches~\cite{sekar2020planning}.

\end{enumerate}

\subsection{Limitations} \label{subsec:model-failures}
The main limitation of our approach surrounds the alignment of the symbolic predictions
$
p_\phi(s_t \mid z_t,h_t)
$
with the ground-truth properties they are meant to represent.
If the predicted symbolic distribution is misaligned, then intervening on the reward function can assign rewards according to the wrong symbolic interpretation, leading to adaptation failure.
Symbolic supervision aligns the symbols with the intended semantics and facilitates reliable task transfer, while purely reward-based training may lead to symbolic shortcuts sufficient only for the training task. That is, \nesywms make explicit the relation between the symbolic information provided during learning and the range of reward interventions that can be trusted at test time.
Purely neural world models do not provide this possibility because they have no interpretable interface for directly specifying new reward functions.

A separate limitation concerns the availability and sufficiency of the symbolic language. Our formulation
assumes that training and test rewards are defined over the same symbolic properties. If a new task
depends on properties outside this vocabulary, the latent RSSM state may still contain useful predictive
information, but provides no interpretable mechanism for wiring the new reward function on top
of it. Transfer is therefore limited to tasks whose rewards can be expressed sufficiently in the chosen
symbolic vocabulary.

\section{Experimental Evaluation}
\label{sec: experiments}
In this section, we propose experiments to test the following three main claims \textbf{(C)} of our work.
\begin{enumerate}[
    label=\textbf{(C\arabic*)},
    ref=C\arabic*,
    leftmargin=0pt,
    labelsep=0.5em,
    labelwidth=0pt,
    itemindent=*,
    align=left,
    itemsep=0.75em
]
\item \label{C1} \textbf{Sample-efficient training.}
We expect NeSy-WMs to improve  sample efficiency with respect to DreamerV3 due to the provided structure of the reward/continue predictors, especially when some symbolic supervision is provided.
\item \label{C2} \textbf{Zero-shot adaptation by imagination planning.}
We expect NeSy-WMs to enable planning on new test tasks under the same dynamics without further learning, unlike existing finetuning approaches.
\item \label{C3} \textbf{Zero-shot adaptation by imagination finetuning.}
We expect NeSy-WMs to enable zero-shot finetuning on novel test tasks completely in the model imagination. This ability contrasts with existing adaptation settings of DreamerV3, which either require online test-task experience or replay buffer relabeling~\cite{sekar2020planning}.
\end{enumerate}
To test our claims, we propose a set of corresponding experiments \textbf{(E)}:
\begin{enumerate}[
    label=\textbf{(E\arabic*)},
    ref=E\arabic*,
    leftmargin=0pt,
    labelsep=0.5em,
    labelwidth=0pt,
    itemindent=*,
    align=left,
    itemsep=0.75em
]

\item \label{E1} \textbf{Training sample efficiency.}
We train \nesywms in fully supervised (\ref{S1}), partially supervised (\ref{S2}), and unsupervised (\ref{S3}) settings to assess their training sample efficiency compared to standard DreamerV3.

\item \label{E2} \textbf{Zero-shot adaptation by planning.}
The trained \nesywms from \ref{E1} are adapted to a variety of test tasks of increasing difficulty \emph{without any further training} through MCTS planning in  model imagination (\ref{O1}). Since Dreamer adaptation requires further training, it cannot be compared in this setting. Instead, we compare against the same MCTS planning algorithm using a perfect environment simulator.

\item \label{E3} \textbf{Zero-shot adaptation by imagination finetuning.}
Using the models trained in \ref{E1}, \nesywms are adapted to a selected set of hard test tasks through imagination finetuning (\ref{O2}). Specifically, \nesywms always freeze their latent dynamics model and train a new actor-critic if symbols were fully supervised (\ref{S1}) or partially supervised (\ref{S2}) during training (\textsc{new AC}). If symbols were unsupervised during training (\ref{S3}), \nesywms instead learn a new actor-critic and reward/continue predictors through replay buffer relabeling following~\citet{sekar2020planning}. For DreamerV3, we employ the same adaptation scheme by training a new actor-critic and reward/continue predictors from the relabeled buffer (\textsc{new AC + R + C}). We also evaluate a baseline where the entire DreamerV3 world model is unfrozen (\textsc{WM unfrozen}).
\end{enumerate}

\begin{figure*}[t]
    \centering
    \includegraphics[width=0.7\linewidth]{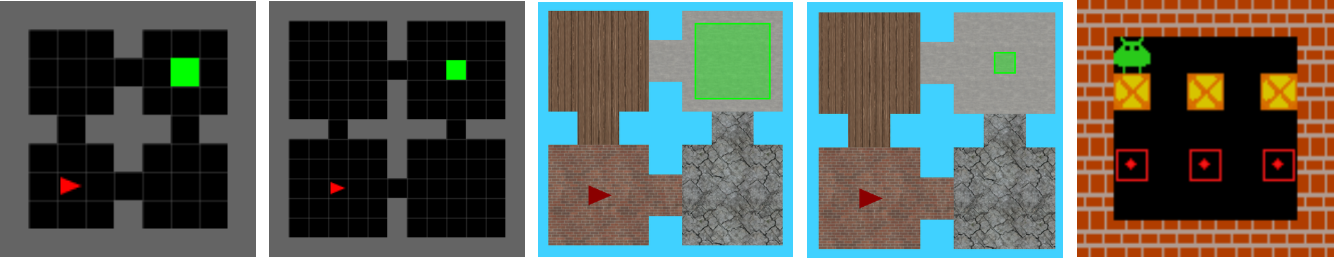}
    \caption{Training tasks. The initial position of the agent is random. From left to right: \textsc{MiniGrid-small}, \textsc{MiniGrid-large}, a top view of \textsc{MiniWorld-small} and \textsc{MiniWorld-large} differing in movement granularity and goal area size, and \textsc{Sokoban}.}
    \label{fig:training-envs}
\end{figure*}
\begin{figure*}[t]
    \centering
    \includegraphics[width=0.8\textwidth]{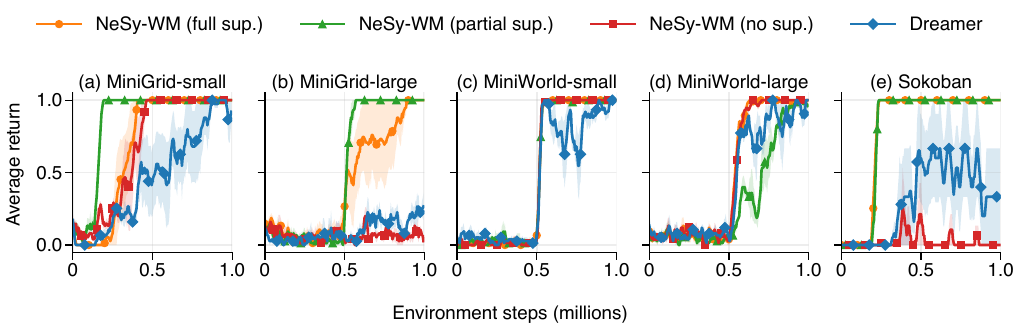}
    \caption{
        Training results. Curves show the average return across three seeds and their standard error. Both NeSy-WMs and Dreamer start by an exploration phase in the \textsc{MiniWorld} environments and in \textsc{MiniGrid-large} (see supplementary material).
    }
    \label{fig:main-training}
\end{figure*}
\begin{table*}[t]
  \centering
  \small
  \setlength{\tabcolsep}{1mm}

  \begin{tabular}{@{}lcccccccccc@{}}
    \toprule
    & \multicolumn{2}{c}{\textsc{MG-small}}
    & \multicolumn{2}{c}{\textsc{MG-large}}
    & \multicolumn{2}{c}{\textsc{MW-small}}
    & \multicolumn{2}{c}{\textsc{MW-large}}
    & \multicolumn{2}{c}{\textsc{Sokoban}} \\
    \cmidrule(lr){2-3}
    \cmidrule(lr){4-5}
    \cmidrule(lr){6-7}
    \cmidrule(lr){8-9}
    \cmidrule(lr){10-11}

    Challenge
    & Full & Partial
    & Full & Partial
    & Full & Partial 
    & Full & Partial
    & Full  & Partial \\
    \midrule

    \textsc{Easy}
    & \mctsresult{1.00}{0.00}{0.0}
    & \mctsresult{1.00}{0.00}{2.7}
    & \mctsresult{1.00}{0.00}{\phantom{0}0.0}
    & \mctsresult{1.00}{0.00}{\phantom{0}0.0}
    & \mctsresult{0.78}{0.15}{\phantom{0}8.7}
    & \mctsresult{0.89}{0.11}{\phantom{0}7.4}
    & \mctsresult{0.89}{0.11}{\phantom{0}5.1}
    & \mctsresult{0.44}{0.18}{24.5}
    & \mctsresult{1.00}{0.00}{\phantom{0}0.0}
    & \mctsresult{1.00}{0.00}{\phantom{0}1.1} \\

    \textsc{Medium}
    & \mctsresult{1.00}{0.00}{0.0}
    & \mctsresult{0.67}{0.17}{0.0}
    & \mctsresult{1.00}{0.00}{\phantom{0}0.0}
    & \mctsresult{1.00}{0.00}{\phantom{0}8.2}
    & \mctsresult{1.00}{0.00}{\phantom{0}8.9}
    & \mctsresult{1.00}{0.00}{\phantom{0}8.2}
    & \mctsresult{0.11}{0.11}{33.0}
    & \mctsnosuccess{0.00}{0.00}
    & \mctsresult{1.00}{0.00}{\phantom{0}1.8}
    & \mctsresult{1.00}{0.00}{\phantom{0}0.2} \\

    \textsc{Hard}
    & \mctsresult{1.00}{0.00}{9.8}
    & \mctsresult{0.67}{0.17}{9.0}
    & \mctsresult{0.22}{0.15}{33.0}
    & \mctsresult{0.44}{0.18}{20.0}
    & \mctsresult{0.67}{0.17}{35.3}
    & \mctsresult{0.67}{0.17}{16.5}
    & \mctsnosuccess{0.00}{0.00}
    & \mctsnosuccess{0.00}{0.00}
    & \mctsresult{0.67}{0.17}{72.5}
    & \mctsresult{0.67}{0.17}{21.8} \\

    \bottomrule
  \end{tabular}
    \caption{
    MCTS evaluation of fully and partially supervised NeSy-WMs. MG and MW denote MiniGrid and MiniWorld, respectively.
    We report success rate $\pm$ standard error over nine episodes
(three episodes for each of three trained models) and the optimality
gap $\Delta=L-L^\star$, where $L$ is the mean length of successful
plans and $L^\star$ is the optimal plan length for that challenge.
  }   \label{tab:mcts-supervision-comparison}
\end{table*}
\begin{figure*}[t]
    \centering
    \includegraphics[width=0.8\textwidth]{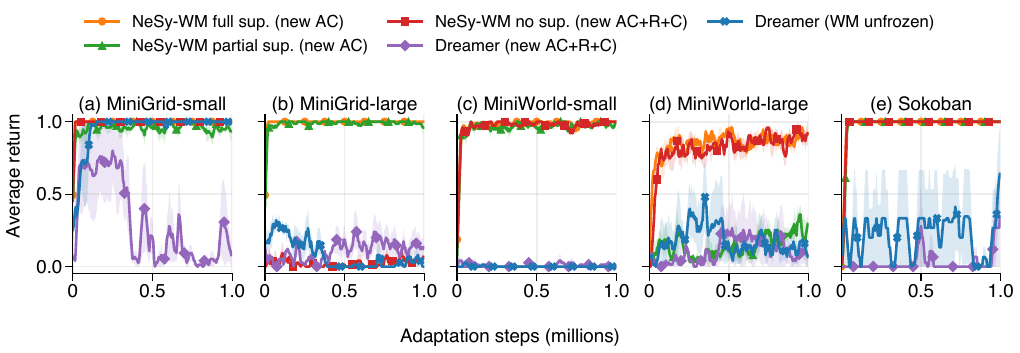}
    \caption{
        Imagination finetuning results. Curves show the average return across three seeds and their standard error. 
        Adaptation is entirely offline; during finetuning, we periodically
evaluate the policy in the test environment solely to measure
performance.
    }
    \label{fig:finetuning}
\end{figure*}
\subsection{Training Environments} 
\label{subsec:exp-setup}
We describe the environments used for the training experiments in \ref{E1} (Figure \ref{fig:training-envs}), and their symbolic abstractions. 
\paragraph{MiniGrid 2D-FourRooms.}
We employ two versions of a 2D Four-Rooms environment implemented using MiniGrid~\cite{MinigridMiniworld23}. In particular, \textsc{MiniGrid-small} has four $3\times3$ rooms, whereas \textsc{MiniGrid-large} has $5\times 5$ rooms. 
The observations are images of the whole grid, and the agent has 3 available actions: \textsc{forward}, \textsc{turn-left}, and \textsc{turn-right}.
The agent only receives a positive reward of +1 if the goal is reached and $0$ otherwise.
At training time, the initial agent position is randomized, and the goal position is fixed as the center of the top-right room.
The $(x, y)$ coordinate of the agent is used as the task-relevant symbolic abstraction.
This abstraction is sufficient for reward prediction, but incomplete for control since the orientation of the agent also matters for choosing the correct action. In experiments with partial symbolic supervision, we only supervise the center of each room.

\paragraph{MiniWorld 3D-FourRooms.}

We employ two versions of a 3D partially observable four-rooms environment with continuous state space, implemented in MiniWorld~\cite{MinigridMiniworld23}. Figure~\ref{fig:training-envs} shows a top-down view (the egocentric view is provided in the supplementary material). The action space matches MiniGrid, except that \textsc{forward} moves the agent by \(0.3\) and \(0.15\) world units in \textsc{MiniWorld-small} and \textsc{MiniWorld-large}, respectively, while turning rotates the agent by \(30^\circ\) and \(15^\circ\). A reward of \(+1\) is received only when the agent enters a goal region. During training, the initial position is randomized and the goal is fixed at the center of the top-right room. We use the continuous \((x,y)\) position as the symbolic abstraction. 
 In experiments with partial symbolic supervision, we only supervise the four potential goal areas centered in each room.
\paragraph{Sokoban.}
Here, the agent needs to push all the boxes to the indicated goal positions in a customized version of the game Sokoban~\cite{SchraderSokoban2018}.
In this version, \emph{it only receives a reward ($+1$) when the task is completed.}
The agent has 9 available actions: \textsc{move-dir} and \textsc{push-dir} for each direction \textsc{dir $\in$ \{left, right, up, down\}}, and a \textsc{null} action which does nothing.  
We employ a task-relevant symbolic abstraction comprising \emph{only the coordinates of the boxes},
since the reward depends only on their locations. In experiments with partial symbolic supervision, we only provide supervision when all boxes are on training or test targets.  

\subsection{Test Tasks}
\paragraph{Planning Tasks} For the experiments in \ref{E2}, we select a set of test tasks for each training environment, visualized in the supplementary material.
In particular, we take the center of the bottom-left room as test goal, opposite the training goal, for MiniGrid and MiniWorld, and move the boxes to opposite (top) target positions in Sokoban. 
The tasks are crafted to identify at what point the planning-adapted model fails as the distance between the initial position of the agent and the goal increases. Specifically, the \textsc{Easy}, \textsc{Medium} and \textsc{Hard} challenges for MiniGrid and MiniWorld feature an increasingly distant starting position of the agent from the goal position. Similarly, the \textsc{Easy} task in \textsc{Sokoban} requires placing only one box in a correct test-task configuration, while \textsc{Hard} requires moving all boxes to new target positions.

\paragraph{Imagination Finetuning Tasks} For the experiments in \ref{E3}, we consider again the opposite map position with respect to training as the goal for MiniGrid and MiniWorld, and moving all three boxes to opposite target positions in Sokoban.
While finetuning the model, we take snapshots of the policy to evaluate it on the test environment with randomized agent initial position at each evaluation episode.

\subsection{Results}
\label{sec:results}

\textbf{(\ref{C1}) \nesywms improve the stability and sample efficiency of world-model training in sparse-reward environments, particularly under symbolic supervision.} Figure~\ref{fig:main-training} shows the training performance of NeSy-WMs under different supervision regimes \ref{S1}-\ref{S3}, compared to DreamerV3. 
Across the five training tasks, NeSy-WMs with either full or partial supervision mostly outperform the baseline in terms of sample-efficiency and stability.
The only exception is the partially supervised agent in \textsc{MiniWorld-large} that converged some steps later.
Supervised NeSy-WMs also converge significantly faster in \textsc{MiniGrid-small}, and outperform the baseline in \textsc{MiniGrid-large} and \textsc{Sokoban}, where Dreamer is unstable and does not converge.
In the case where no symbolic supervision is provided, NeSy-WMs still perform comparably to Dreamer, outperforming it on \textsc{MiniGrid-small} but matching Dreamer's failures on \textsc{Sokoban} and \textsc{MiniGrid-large}.
The benefits provided by the supervision in these experiments suggest that symbol prediction may by itself be a strong learning signal for RL world models.
We further investigated this hypothesis in an ablation reported in the supplementary material, where we show that our performance under symbolic supervision remains equally strong without reconstruction gradients, unlike Dreamer which dramatically fails.

\textbf{(\ref{C2}) NeSy-WMs enable zero-shot MCTS test-time adaptation, but are constrained by the limitations of uninformed planning.} Table~\ref{tab:mcts-supervision-comparison} shows the results of the zero-shot MCTS planning with fully supervised and partially supervised NeSy-WMs on the selected test tasks. 
With the same planning budget of 256 tree expansions per step, the fully supervised model solves most of the \textsc{Easy} and \textsc{Medium} challenges in an optimal or close-to-optimal number of steps.
However, pure planning is not enough to consistently solve the \textsc{Hard} tasks in \textsc{MiniGrid-large}, \textsc{MiniWorld-small}/\textsc{large}, and Sokoban. The partially supervised model generally achieves comparable performance, despite lower success rates in \textsc{MiniGrid-small}.
These results were compared to planning with perfect environment simulators using the same planning budget and MCTS parameters.
Aggregated across all challenges, fully and partially supervised \nesywms achieve success rates of $75.6\%$ and $69.6\%$, respectively, compared with $79.3\%$ for perfect-simulator planning.
The complete comparison is reported in the supplementary material. While increasing the planning budget may improve the results, doing so with a learned model without
value bootstrapping is slow~\cite{muzero} and prone to dynamics degradation~\cite{talvitie2017self, asadi2019combating}.

\textbf{(\ref{C3}) NeSy-WMs  outperform DreamerV3  in the zero-shot imagination finetuning setting.} Figure~\ref{fig:finetuning} shows the adaptation performance of NeSy-WMs against the two Dreamer adaptation regimes.
We remark that while supervised NeSy-WMs only need to train a fresh actor-critic in imagination \textit{without any replay buffer relabeling}, both Dreamer baselines and  unsupervised NeSy-WMs need to train new reward/continue predictors on the relabeled buffer.
The results clearly show how Dreamer consistently fails to adapt to the test-time tasks, with the only exception being \textsc{MiniGrid-small} for the \textsc{WM unfrozen} regime.
Conversely, fully supervised \nesywms adapt almost immediately to all the test-time tasks, followed by partially supervised \nesywms which only struggle in \textsc{MiniWorld-large}.
Unsupervised NeSy-WMs also result in a substantially more reliable adaptation approach than Dreamer, successfully adapting in four out of the five challenges. This shows that even when symbolic supervision is not available, the structure of the neurosymbolic predictors can still make NeSy-WMs a convenient alternative for efficient task-adaptation by imagination finetuning.

\section{Conclusion}
\label{sec: conclusion}
In this work, we presented Neurosymbolic World Models (NeSy-WMs) that build on top of generative world models by incorporating symbolic reward and continuation predictors over learned, task-relevant symbolic properties of the state.
\nesywms demonstrate a consistently more sample-efficient training than DreamerV3 on the sparse-reward tasks considered in this work. 
They can also be employed for zero-shot test-time planning on different tasks than the one they were trained on, i.e. new reward functions defined over the same set of symbolic properties. 
Finally, \nesywms demonstrated effective zero-shot finetuning on novel tasks in imagination without any interaction with the test environment.
Importantly, they consistently outperformed the common approach of finetuning DreamerV3 with a relabeled replay buffer.
Overall, we showed how prior symbolic knowledge of the reward function can be incorporated into generative world models in a way that is simple, benefits the training process, and provides distinctive generalisation properties. 

Promising directions for extending \nesywms include robust dynamics learning for reliable test-time planning over long horizons, testing other forms of symbolic alignment such as temporal consistency constraints, and exploration strategies that expose the model to the symbolic states needed for reliable task transfer in hard-exploration environments.

\section{Acknowledgments}
This research has received funding from the KU Leuven Research Funds (C14/24/092) and from the Flemish Government under the "Onderzoeksprogramma Artificiële Intelligentie (AI) Vlaanderen" programme.

\bibliography{preprint2027}

@article{hafner2025mastering,
  title={Mastering diverse control tasks through world models},
  author={Hafner, Danijar and Pasukonis, Jurgis and Ba, Jimmy and Lillicrap, Timothy},
  journal={Nature},
  volume={640},
  number={8059},
  pages={647--653},
  year={2025},
  publisher={Nature Publishing Group UK London}
}

@inproceedings{cano2025neurosymbolic,
  title={Neurosymbolic World Models for Sequential Decision Making},
  author={Cano, Leonardo Hernandez and Perroni-Scharf, Maxine and Dhir, Neil and Ramamurthy, Arun and Solar-Lezama, Armando},
  booktitle={Forty-second International Conference on Machine Learning},
  year={2025}
}

@inproceedings{sehgalneurosymbolic,
  title={Neurosymbolic Grounding for Compositional World Models},
  author={Sehgal, Atharva and Grayeli, Arya and Sun, Jennifer J and Chaudhuri, Swarat},
  booktitle={The Twelfth International Conference on Learning Representations},
  year={2024}
}

@inproceedings{schaul2015universal,
  title={Universal value function approximators},
  author={Schaul, Tom and Horgan, Daniel and Gregor, Karol and Silver, David},
  booktitle={International conference on machine learning},
  pages={1312--1320},
  year={2015},
  organization={PMLR}
}

@inproceedings{liu2022goal,
  title={Goal-Conditioned Reinforcement Learning: Problems and Solutions},
  author={Liu, Minghuan and Zhu, Menghui and Zhang, Weinan},
  booktitle={Proceedings of the Thirty-First International Joint Conference on Artificial Intelligence},
  pages={5502--5511},
  year={2022},
  organization={International Joint Conferences on Artificial Intelligence Organization}
}

@article{andrychowicz2017hindsight,
  title={Hindsight experience replay},
  author = {Andrychowicz, Marcin and
          Wolski, Filip and
          Ray, Alex and
          Schneider, Jonas and
          Fong, Rachel and
          Welinder, Peter and
          McGrew, Bob and
          Tobin, Josh and
          Abbeel, Pieter and
          Zaremba, Wojciech},
  journal={Advances in neural information processing systems},
  volume={30},
  year={2017}
}

@article{lambrechts2024informed,
  title={Informed POMDP: Leveraging Additional Information in Model-Based RL},
  author={Lambrechts, Gaspard and Bolland, Adrien and Ernst, Damien},
  journal={Reinforcement Learning Journal},
  year={2024}
}

@inproceedings{huang2025pigdreamer,
  title={PIGDreamer: Privileged information guided world models for safe partially observable reinforcement learning},
author = {Huang, Dongchi and
          Wang, Jiaqi and
          Li, Yang and
          Xia, Chunhe and
          Zhang, Tianle and
          Zhang, Kaige},
  booktitle={Forty-second International Conference on Machine Learning},
  year={2025}
}

@inproceedings{tdmpc2,
  title={TD-MPC2: Scalable, Robust World Models for Continuous Control},
  author={Hansen, Nicklas and Su, Hao and Wang, Xiaolong},
  booktitle={The Twelfth International Conference on Learning Representations},
  year={2024}
}

@inproceedings{tdmpc,
  title={Temporal Difference Learning for Model Predictive Control},
  author={Hansen, N and Wang, X and Su, H},
  booktitle={International Conference on Machine Learning, PMLR},
  year={2022}
}

@inproceedings{hafnerdreamv1,
  title={Dream to Control: Learning Behaviors by Latent Imagination},
  author={Hafner, Danijar and Lillicrap, Timothy and Ba, Jimmy and Norouzi, Mohammad},
  booktitle={International Conference on Learning Representations},
  year={2020}
}

@inproceedings{hafnermasteringv2,
  title={Mastering Atari with Discrete World Models},
  author={Hafner, Danijar and Lillicrap, Timothy P and Norouzi, Mohammad and Ba, Jimmy},
  booktitle={International Conference on Learning Representations},
  year={2021}
}

@article{misino2022vael,
  title={Vael: Bridging variational autoencoders and probabilistic logic programming},
  author={Misino, Eleonora and Marra, Giuseppe and Sansone, Emanuele},
  journal={Advances in Neural Information Processing Systems},
  volume={35},
  pages={4667--4679},
  year={2022}
}

@inproceedings{hafner2019learning,
  title={Learning latent dynamics for planning from pixels},
  author={Hafner, Danijar and Lillicrap, Timothy and Fischer, Ian and Villegas, Ruben and Ha, David and Lee, Honglak and Davidson, James},
  booktitle={International conference on machine learning},
  pages={2555--2565},
  year={2019},
  organization={PMLR}
}

@inproceedings{talvitie2017self,
  title={Self-correcting models for model-based reinforcement learning},
  author={Talvitie, Erik},
  booktitle={Proceedings of the AAAI conference on artificial intelligence},
  volume={31},
  year={2017}
}

@article{asadi2019combating,
  title={Combating the compounding-error problem with a multi-step model},
  author={Asadi, Kavosh and Misra, Dipendra and Kim, Seungchan and Littman, Michel L},
  journal={arXiv preprint arXiv:1905.13320},
  year={2019}
}

@inproceedings{sekar2020planning,
  title={Planning to explore via self-supervised world models},
  author={Sekar, Ramanan and Rybkin, Oleh and Daniilidis, Kostas and Abbeel, Pieter and Hafner, Danijar and Pathak, Deepak},
  booktitle={International conference on machine learning},
  pages={8583--8592},
  year={2020},
  organization={PMLR}
}

@article{ha2018recurrent,
  title={Recurrent world models facilitate policy evolution},
  author={Ha, David and Schmidhuber, J{\"u}rgen},
  journal={Advances in neural information processing systems},
  volume={31},
  year={2018}
}

@inproceedings{micheli2023transformers,
  title={Transformers are Sample-Efficient World Models},
  author={Micheli, Vincent and Alonso, Eloi and Fleuret, Fran{\c{c}}ois},
  booktitle={The Eleventh International Conference on Learning Representations},
  year={2023}
}

@inproceedings{desmet2025relational,
  title={Relational neurosymbolic Markov models},
  author={De Smet, Lennert and Venturato, Gabriele and De Raedt, Luc and Marra, Giuseppe},
  booktitle={Proceedings of the AAAI Conference on Artificial Intelligence},
  volume={39},
  pages={16181--16189},
  year={2025}
}

@article{paszke2019pytorch,
  title={Pytorch: An imperative style, high-performance deep learning library},
  author={Paszke, Adam and Gross, Sam and Massa, Francisco and Lerer, Adam and Bradbury, James and Chanan, Gregory and Killeen, Trevor and Lin, Zeming and Gimelshein, Natalia and Antiga, Luca and others},
  journal={Advances in neural information processing systems},
  volume={32},
  year={2019}
}

@inproceedings{danihelka2022policy,
  title={Policy improvement by planning with Gumbel},
  author={Danihelka, Ivo and Guez, Arthur and Schrittwieser, Julian and Silver, David},
  booktitle={International Conference on Learning Representations},
  year={2022}
}

@inproceedings{zhou2025dino,
  title={DINO-WM: World Models on Pre-trained Visual Features enable Zero-shot Planning},
  author={Zhou, Gaoyue and Pan, Hengkai and LeCun, Yann and Pinto, Lerrel},
  booktitle={International Conference on Machine Learning},
  pages={79115--79135},
  year={2025},
  organization={PMLR}
}

@inproceedings{assran2023self,
  title={Self-supervised learning from images with a joint-embedding predictive architecture},
  author={Assran, Mahmoud and Duval, Quentin and Misra, Ishan and Bojanowski, Piotr and Vincent, Pascal and Rabbat, Michael and LeCun, Yann and Ballas, Nicolas},
  booktitle={Proceedings of the IEEE/CVF conference on computer vision and pattern recognition},
  pages={15619--15629},
  year={2023}
}

@inproceedings{de2023neural,
  title={Neural probabilistic logic programming in discrete-continuous domains},
  author={De Smet, Lennert and Dos Martires, Pedro Zuidberg and Manhaeve, Robin and Marra, Giuseppe and Kimmig, Angelika and De Raedt, Luc},
  booktitle={Uncertainty in Artificial Intelligence},
  pages={529--538},
  year={2023},
  organization={PMLR}
}

@inproceedings{yang2020neurasp,
  title={NeurASP: Embracing neural networks into answer set programming},
  author={Yang, Zhun and Ishay, Adam and Lee, Joohyung},
  booktitle={29th International Joint Conference on Artificial Intelligence, IJCAI 2020},
  pages={1755--1762},
  year={2020},
  organization={International Joint Conferences on Artificial Intelligence}
}

@inproceedings{zeng2020scaling,
  title={Scaling up hybrid probabilistic inference with logical and arithmetic constraints via message passing},
  author={Zeng, Zhe and Morettin, Paolo and Yan, Fanqi and Vergari, Antonio and Van den Broeck, Guy},
  booktitle={International Conference on Machine Learning},
  pages={10990--11000},
  year={2020},
  organization={PMLR}
}

@inproceedings{maene2025klay,
  title={KLay: Accelerating Arithmetic Circuits for Neurosymbolic AI},
  author={Maene, Jaron and Derkinderen, Vincent and Zuidberg Dos Martires, Pedro},
  booktitle={The Thirteenth International Conference on Learning Representations},
  year={2025},
}

@article{roth1996hardness,
  title={On the hardness of approximate reasoning},
  author={Roth, Dan},
  journal={Artificial intelligence},
  volume={82},
  number={1-2},
  pages={273--302},
  year={1996},
  publisher={Elsevier}
}

@article{chavira2008probabilistic,
  title={On probabilistic inference by weighted model counting},
  author={Chavira, Mark and Darwiche, Adnan},
  journal={Artificial Intelligence},
  volume={172},
  number={6-7},
  pages={772--799},
  year={2008},
  publisher={Elsevier}
}

@inproceedings{belle2015probabilistic,
  title={Probabilistic inference in hybrid domains by weighted model integration},
  author={Belle, Vaishak and Passerini, Andrea and Van den Broeck, Guy},
  booktitle={Proceedings of the Twenty-Fourth International Joint Conference on Artificial Intelligence, IJCAI 2015, Buenos Aires, Argentina, July 25-31, 2015},
  pages={2770--2776},
  year={2015},
  organization={IJCAI Inc}
}

@article{garcez2023neurosymbolic,
  title={Neurosymbolic AI: The 3rd wave},
  author={Garcez, Artur d’Avila and Lamb, Luis C},
  journal={Artificial Intelligence Review},
  volume={56},
  number={11},
  pages={12387--12406},
  year={2023},
  publisher={Springer}
}

@article{marra2024statistical,
  title={From statistical relational to neurosymbolic artificial intelligence: A survey},
  author={Marra, Giuseppe and Duman{\v{c}}i{\'c}, Sebastijan and Manhaeve, Robin and De Raedt, Luc},
  journal={Artificial Intelligence},
  volume={328},
  pages={104062},
  year={2024},
  publisher={Elsevier}
}

@inproceedings{kingma2014auto,
  author       = {Diederik P. Kingma and
                  Max Welling},
  title        = {Auto-Encoding Variational Bayes},
  booktitle    = {2nd International Conference on Learning Representations {ICLR}},
  year         = {2014},
}

@article{alonso2024diffusion,
  title={Diffusion for world modeling: Visual details matter in atari},
  author={Alonso, Eloi and Jelley, Adam and Micheli, Vincent and Kanervisto, Anssi and Storkey, Amos and Pearce, Tim and Fleuret, Fran{\c{c}}ois},
  journal={Advances in Neural Information Processing Systems},
  volume={37},
  pages={58757--58791},
  year={2024}
}

@article{rubinstein1997optimization,
  title={Optimization of computer simulation models with rare events},
  author={Rubinstein, Reuven Y},
  journal={European Journal of Operational Research},
  volume={99},
  number={1},
  pages={89--112},
  year={1997},
  publisher={Elsevier}
}

@inproceedings{coulom2006efficient,
  title={Efficient selectivity and backup operators in Monte-Carlo tree search},
  author={Coulom, R{\'e}mi},
  booktitle={International conference on computers and games},
  pages={72--83},
  year={2006},
  organization={Springer}
}

@inproceedings{kocsis2006bandit,
  title={Bandit Based Monte-Carlo Planning},
  author={Kocsis, Levente and Szepesv{\'a}ri, Csaba},
  booktitle={European conference on machine learning},
  pages={282--293},
  year={2006},
  organization={Springer}
}

@article{icarte2022reward,
  title={Reward machines: Exploiting reward function structure in reinforcement learning},
  author={Icarte, Rodrigo Toro and Klassen, Toryn Q and Valenzano, Richard and McIlraith, Sheila A},
  journal={Journal of Artificial Intelligence Research},
  volume={73},
  pages={173--208},
  year={2022}
}

@article{MinigridMiniworld23,
  author       = {Maxime Chevalier-Boisvert and Bolun Dai and Mark Towers and Rodrigo de Lazcano and Lucas Willems and Salem Lahlou and Suman Pal and Pablo Samuel Castro and Jordan Terry},
  title        = {Minigrid \& Miniworld: Modular \& Customizable Reinforcement Learning Environments for Goal-Oriented Tasks},
  journal      = {CoRR},
  volume       = {abs/2306.13831},
  year         = {2023},
}

@misc{SchraderSokoban2018,
  author = {Schrader, Max-Philipp B.},
  title = {gym-sokoban},
  year = {2018},
  publisher = {GitHub},
  journal = {GitHub repository},
  howpublished = {\url{https://github.com/mpSchrader/gym-sokoban}},
}

@phdthesis{couetoux2013monte,
  title={Monte Carlo tree search for continuous and stochastic sequential decision making problems},
  author={Couetoux, Adrien},
  year={2013},
  school={Universit{\'e} Paris Sud-Paris XI}
}

@inproceedings{marconato2024bears,
  title={BEARS Make Neuro-Symbolic Models Aware of their Reasoning Shortcuts},
  author={Marconato, Emanuele and Bortolotti, Samuele and van Krieken, Emile and Vergari, Antonio and Passerini, Andrea and Teso, Stefano},
  booktitle={Uncertainty in Artificial Intelligence},
  pages={2399--2433},
  year={2024},
  organization={PMLR}
}

@article{marconato2023not,
  title={Not all neuro-symbolic concepts are created equal: Analysis and mitigation of reasoning shortcuts},
  author={Marconato, Emanuele and Teso, Stefano and Vergari, Antonio and Passerini, Andrea},
  journal={Advances in Neural Information Processing Systems},
  volume={36},
  pages={72507--72539},
  year={2023}
}

@article{manhaeve2018deepproblog,
  title={Deepproblog: Neural probabilistic logic programming},
  author={Manhaeve, Robin and Dumancic, Sebastijan and Kimmig, Angelika and Demeester, Thomas and De Raedt, Luc},
  journal={Advances in neural information processing systems},
  volume={31},
  year={2018}
}

@article{muzero,
  title={Mastering atari, go, chess and shogi by planning with a learned model},
  author={Schrittwieser, Julian and Antonoglou, Ioannis and Hubert, Thomas and Simonyan, Karen and Sifre, Laurent and Schmitt, Simon and Guez, Arthur and Lockhart, Edward and Hassabis, Demis and Graepel, Thore and others},
  journal={Nature},
  volume={588},
  number={7839},
  pages={604--609},
  year={2020},
  publisher={Nature Publishing Group UK London}
}

\end{document}